\documentclass[11pt,a4paper]{article}
\usepackage[hyperref]{acl2019}
\usepackage{times}
\usepackage{latexsym}

\usepackage{url}
\usepackage{helvet}
\usepackage{courier}
\usepackage{url}
\usepackage{graphicx}
\usepackage[english]{babel}
\usepackage{amsmath}
\usepackage{amssymb}
\usepackage{mathtools}
\usepackage{color}
\usepackage{multirow}

\aclfinalcopy
\DeclareMathSymbol{\mlq}{\mathord}{operators}{``}
\DeclareMathSymbol{\mrq}{\mathord}{operators}{`'}
\newcommand\bigforall{\mbox{\Large $\mathsurround0pt\forall$}} 
\newcommand{\textunderscript}[1]{$_{\text{#1}}$}

\newcommand{\ie}[1]{\emph{i.e.}}
\newcommand{\eg}[1]{\emph{e.g.}}

\newcommand{\qex}[1]{`{\em #1}'}
\newcommand{\mathqex}[1]{\mlq\mathit{#1}\mrq}

\newcommand{\tagset}[0]{tag-set}
\newcommand{\Tagset}[0]{Tag-set}
\newcommand{\tagsets}[0]{tag-sets}
\newcommand{\Tagsets}[0]{Tag-sets}
\newcommand{\SOTA}[0]{state-of-the-art}
\newcommand{\HETTAGS}[0]{heterogeneous \tagsets{}}

\newcommand{\HETTAG}[0]{heterogeneous \tagset{}}
\newcommand{\NONE}[0]{Other}
\newcommand{\NONEmath}[0]{\emph{\NONE}}
\newcommand{\FGTSmath}[0]{T^{FG}}
\newcommand{\FGTS}[0]{$\FGTSmath$}
\newcommand{\ISIX}[0]{\textbf{\emph{I2B2'06}}}
\newcommand{\IFOURTEEN}[0]{\textbf{\emph{I2B2'14}}}
\newcommand{\PHYS}[0]{\textbf{\emph{Physio}}}
\newcommand{\ONTO}[0]{\textbf{\emph{Onto}}}
\newcommand{\CONLL}[0]{\textbf{\emph{Conll}}}

\newcommand{\IndM}[0]{\textbf{\emph{M\textunderscript{Indep}}}}
\newcommand{\MtlM}[0]{\textbf{\emph{M\textunderscript{MTL}}}}
\newcommand{\HierM}[0]{\textbf{\emph{M\textunderscript{Hier}}}}
\newcommand{\ConcatM}[0]{\textbf{\emph{M\textunderscript{Concat}}}}

\title{A Joint Named-Entity Recognizer for Heterogeneous \Tagsets{}\\ Using a Tag Hierarchy}
\author{Genady Beryozkin, Yoel Drori, Oren Gilon, Tzvika Hartman and Idan Szpektor \\
Google Research\\
  Tel Aviv, Israel \\
  {\tt \{genady,dyoel,ogilon,tzvika,szpektor\} @google.com}}

\date{}

\begin{document}

\maketitle

\begin{abstract}
We study a variant of domain adaptation for named-entity recognition where multiple, heterogeneously tagged training sets are available. Furthermore, 
the test \tagset{} is not identical to any individual training \tagset{}. Yet, the relations between all tags are provided in a tag hierarchy, covering the test tags as a combination of training tags.
This setting occurs when various datasets are created using different annotation schemes. 
This is also the case of extending a \tagset{} with a new tag by annotating only the new tag in a new dataset.
We propose to use the given tag hierarchy to jointly learn a neural network that shares its tagging layer among all \tagsets{}.
We compare this model to combining independent models and to a model based on the multitasking approach.
Our experiments show the benefit of the tag-hierarchy model, especially when facing non-trivial consolidation of \tagsets{}.

\end{abstract}

\section{Introduction}
\label{sec:intro}








Named Entity Recognition (NER) has seen significant progress in the last couple of years with the application of Neural Networks to the task. Such models achieve \SOTA{} performance with little or no manual feature engineering \cite{Collobert2011,Huang2015NNbasedNER,Lample2016NNbasedNER,Ma2016,Dernoncourt2017}. Following this success, more complex NER setups are approached with neural models, among them domain adaptation \cite{Qu2016,He2017,Dong2017}. 

\begin{figure}[t]
\centering
\scriptsize
\includegraphics[width=\linewidth]{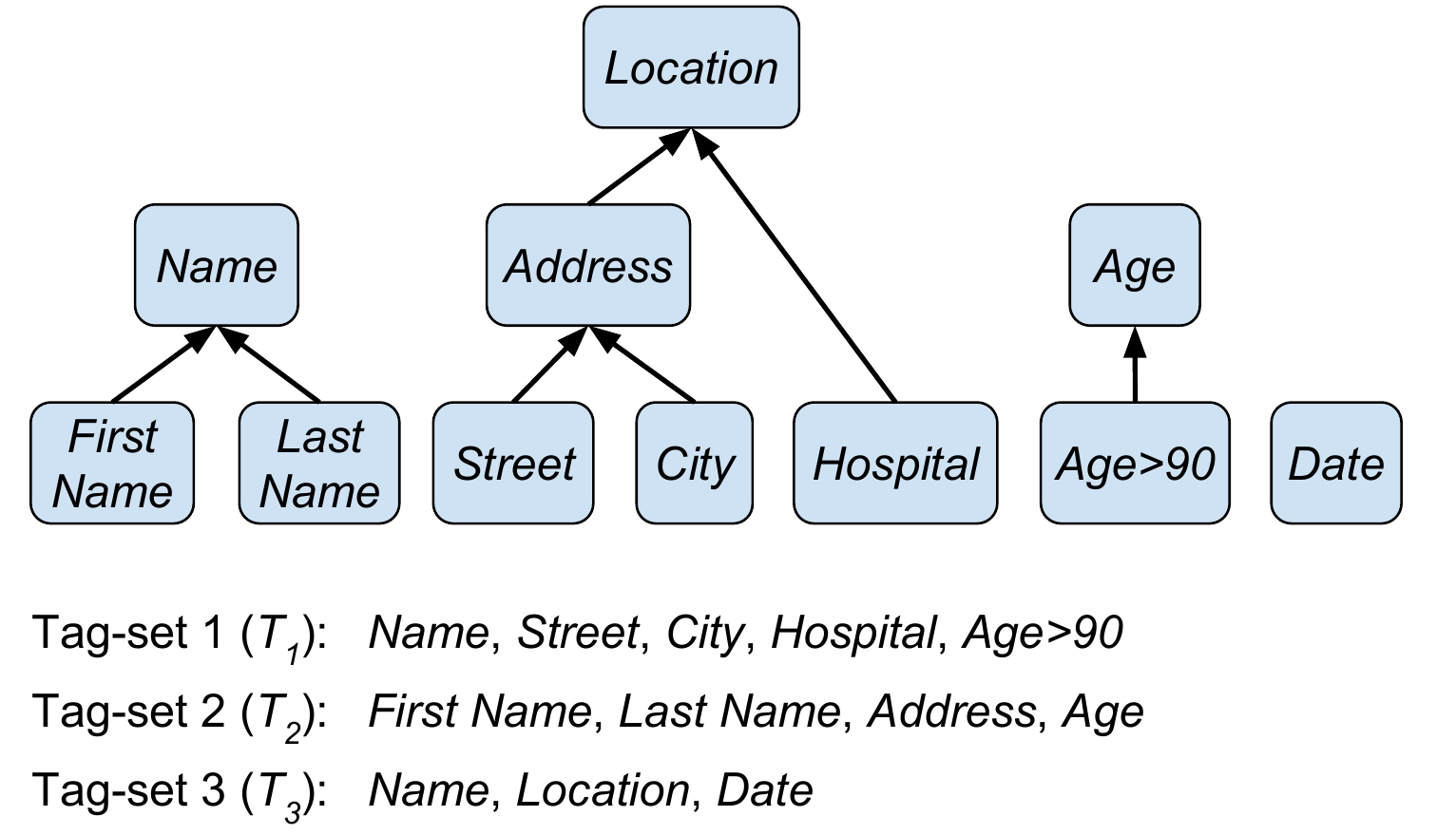}
\caption{A tag hierarchy for three \tagsets{}.}
\label{fig:hierarchy}
\end{figure}

In this work we study one type of domain adaptation for NER, denoted here \emph{\HETTAGS{}}. In this variant, samples from the test set are not available at training time. Furthermore, the test \tagset{} differs from each training \tagset{}. However every test tag can be represented either as a single training tag or as a combination of several training tags. This information is given in the form of a hypernym hierarchy over all tags, training and test (see Fig.~\ref{fig:hierarchy}).


This setting arises when different schemes are used for annotating multiple datasets for the same task.
This often occurs in the medical domain, where healthcare providers use customized \tagsets{} to create their own private test sets \cite{Shickel2017,Lee2017}.
Another scenario is selective annotation, as in the case of extending an existing \tagset{}, \eg{} \{\qex{Name}, \qex{Location}\}, with another tag, \eg{} \qex{Date}. To save annotation effort, new training data is labeled only with the new tag. This case of disjoint \tagsets{} is also discussed in the work of \citet{greenberg2018}. A similar case is extending a training-set with new examples in which only rare tags are annotated. 
In domains where training data is scarce, out-of-domain datasets annotated with infrequent tags may be very valuable.

A naive approach concatenates all training-sets, ignoring the differences between the tagging schemes in each example. A different approach would be to learn to tag with multiple training \tagsets{}. Then, in a post-processing step, the predictions from the different \tagsets{} need to be consolidated into a single test tag sequence, resolving tagging differences along the way. We study two such models. The first model learns an independent NER model for each training \tagset{}. The second model applies the multitasking (MTL) \cite{Collobert2011,Ruder2017} paradigm, in which a shared latent representation of the input text is fed into separate tagging layers.





The above models require heuristic post-processing to consolidate the different predicted tag sequences. To overcome this limitation, we propose a model that incorporates the given tag hierarchy within the neural NER model. Specifically, this model learns to predict a tag sequence only over the fine-grained tags in the hierarchy. At training time, gradients on each dataset-specific labeled examples are propagated as gradients on plausible fine-grained tags. At inference time the model predicts a single sequence of fine-grained tags, which are then mapped to the test \tagset{} by traversing the tag hierarchy. Importantly, all tagging decisions are performed in the model without the need for a post-processing consolidation step.


We conducted two experiments. The first evaluated the extension of a \tagset{} with a new tag via selective annotation of a new dataset with only the extending tag, using datasets from the medical and news domains. In the second experiment we integrated two full \tagsets{} from the medical domain with their training data while evaluating on a third test \tagset{}.
The results show that the model which incorporates the tag-hierarchy is more robust compared to a combination of independent models or MTL, and typically outperforms them. This is especially evident when many tagging collisions need to be settled at post-processing. In these cases, the performance gap in favor of the tag-hierarchy model is large. 

\section{Background and Definitions}
\subsection{Task Definition}
\label{sec:task}

The goal in the \textit{\HETTAGS{}} domain adaptation task is to learn an NER model $M$ that given an input token sequence $x=\{x_i\}_1^n$ infers a tag sequence $y\!=\!\{y_i\}_1^n\!=\!M(x)$ over a test \tagset{} $T^s$, $\forall_i\,y_i\!\!\!\in\!\!\!T^s$. 
To learn the model, $K$ training datasets $\{DS^r_k\}_{k=1}^K$ are provided, each labeled with its own \tagset{} $T^r_k$.
Superscripts 's' and 'r' stand for 'te\underline{s}t' and 't\underline{r}aining', respectively.
In this task, no training \tagset{} is identical to the test \tagset{} $T^s$ by itself. However, all tags in $T^s$ can be covered by combining the training \tagsets{} $\{T^r_k\}_{k=1}^K$. This information is provided in the form of a directed acyclic graph (DAG) representing hypernymy relations between all training and test tags. Fig.~\ref{fig:hierarchy} illustrates such a hierarchy.

As mentioned above, an example scenario is selective annotation, in which an original \tagset{} is extended with a new tag $t$, each with its own training data, and the test \tagset{} is their union.
But, some setups require combinations other than a simple union, \eg{} covering the test tag \qex{Address} with the finer training tags \qex{Street} and \qex{City}, each from a different \tagset{}.



This task is different from inductive domain adaptation \cite{Pan2010,Ruder2017}, in which the \tagsets{} are different but the tasks differ as well (\eg{} NER and parsing), with no need to map the outcomes to a single \tagset{} at test time. 

\subsection{Neural network for NER}
\label{ssec:base_model}

\begin{figure}[t]
\centering
\scriptsize
\includegraphics[width=0.72\linewidth]{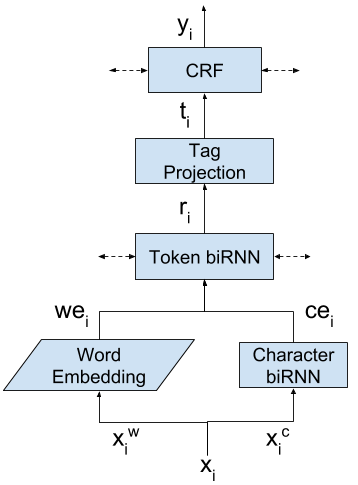}
\caption{Neural architecture for NER.}
\label{fig:LSTMCRF}
\end{figure}

As the underlying architecture shared by all models in this paper, we follow the neural network proposed by \citet{Lample2016NNbasedNER}, which achieved \SOTA{} results on NER. In this model, depicted in Fig.~\ref{fig:LSTMCRF}, each input token $x_i$ is represented as a combination of: (a) a one-hot vector $x^w_i$, mapping the input to a fixed word vocabulary, and (b) a sequence of one-hot vectors $\{x^c_{i,j}\}_{j=1}^{n_i}$, representing the input word's character sequence. 

Each input token $x_i$ is first embedded in latent space by applying both a word-embedding matrix, $we_i = E\,x^w_i$, and a character-based embedding layer $ce_i = \mathrm{CharBiRNN}(\{x^c_{i,j}\})$ \cite{CharBasedEmbeddings}. This output of this step is $e_i = ce_i \oplus we_i$, where $\oplus$ stands for vector concatenation. 
Then, the embedding vector sequence $\{e_i\}_1^n$ is re-encoded in context using a bidirectional RNN layer $\{r_i\}_1^n=\mathrm{BiRNN}(\{e_i\}_1^n)$ \cite{BIRNN}. The sequence $\{r_i\}_1^n$ constitutes the latent representation of the input text.

Finally, each re-encoded vector $r_i$ is projected to tag space for the target \tagset{} $T$, $t_i = P\,r_i$, where $|t_i| = |T|$. 
The sequence $\{t_i\}_1^n$ 
is then taken as input to a CRF layer \cite{CRF}, which maintains a global tag transition matrix. At inference time, the model output is $y=M(x)$, the most probable CRF tag sequence for input $x$.




\section{Models for Multiple Tagging Layers}
\label{sec:mtl_adaptation}

One way to learn a model for the \HETTAGS{} setting is to train a base NER (Sec.~\ref{ssec:base_model}) on the concatenation of all training-sets, predicting tags from the union of all training \tagsets{}. In our experiments, this model under performed, due to the fact that it treats each training example as fully tagged despite being tagged only with the tags belonging to the training-set from which the example is taken (see Sec.~\ref{sec:results}). 

We next present two models that instead learn to tag each training \tagset{} separately. In the first model the outputs from independent base models, each trained on a different \tagset{}, are merged. The second model utilizes the the multitasking approach to train separate tagging layers that share a single text representation layer.



\subsection{Combining independent models}
\label{ssec:base_adaptation}

\begin{figure}
\centering
\includegraphics[width=0.72\linewidth]{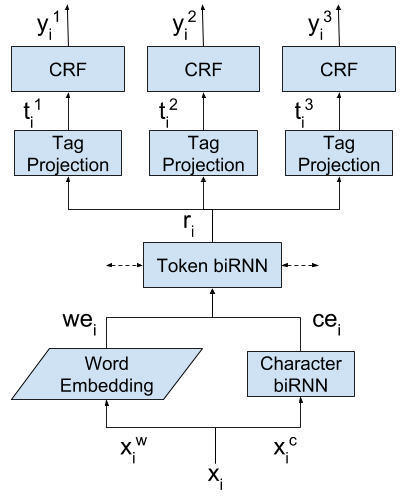}
\caption{NER multitasking architecture for 3 \tagsets{}.}
\label{fig:multitasking}
\end{figure}

In this model, we train a separate NER model for each training set, resulting in $K$ models $\{M_k\}_{k=1}^K$. 
At test time, each model predicts a sequence $y_k=M_k(x)$ over the corresponding \tagset{} $T^r_k$. The sequences $\{y_k\}_{k=1}^K$ are consolidated into a single sequence $y^s$ over the test \tagset{} $T^s$. 

We perform this consolidation in a post-processing step. First, each predicted tag $y_{k,i}$ is mapped to the test \tagset{} as $y^s_{k,i}$. We employ the provided tag hierarchy for this mapping by traversing it starting from $y_{k,i}$ until a test tag is reached. Then, for every token $x_i$, we consider the test tags predicted at position $i$ by the different models $M(x_i)=\{y^s_{k,i} | y^s_{k,i} \ne \mathqex{\NONEmath{}}\}$. Cases where $M(x_i)$ contains more than one tag are called \textit{collisions}. Models must consolidate collisions, selecting a single predicted tag for $x_i$.

We introduce three different consolidation methods. The first is to randomly select a tag from $M(x_i)$.
The second chooses the tag that originates from the tag sequence $y_k$ with the highest CRF probability score.
The third computes the marginal CRF tag probability for each tag and selects the one with the highest probability.



\subsection{Multitasking for \HETTAGS{}}
\label{ssec:multitasking}

Lately, several works explored using multitasking (MTL) for inductive transfer learning within a neural architecture \cite{Collobert2008,Chen2016,Peng2017}. Such algorithms jointly train a single model to solve different NLP tasks, such as NER, sentiment analysis and text classification. The various tasks share the same text representation layer in the model but maintain a separate tagging layer per task.

We adapt multitasking to \HETTAGS{} by considering each training dataset, which has a different \tagset{} $T^r_k$, as a separate NER task. Thus, a single model is trained, in which the latent text representation $\{r_i\}_1^n$ (see Sec.~\ref{ssec:base_model}) is shared between NER tasks. As mentioned above, the tagging layers (projection and CRF) are kept separate for each \tagset{}. Fig.~\ref{fig:multitasking} illustrates this architecture.

We emphasize that the output of the MTL model still consists of $\{y_k\}_{k=1}^K$ different tag sequence predictions. They are consolidated into a final single sequence $y^s$ using the same post-processing step described in Sec.~\ref{ssec:base_adaptation}.

\section{Tag Hierarchy Model}
\label{sec:model}

\begin{figure}
\centering
\scriptsize
\includegraphics[width=\linewidth]{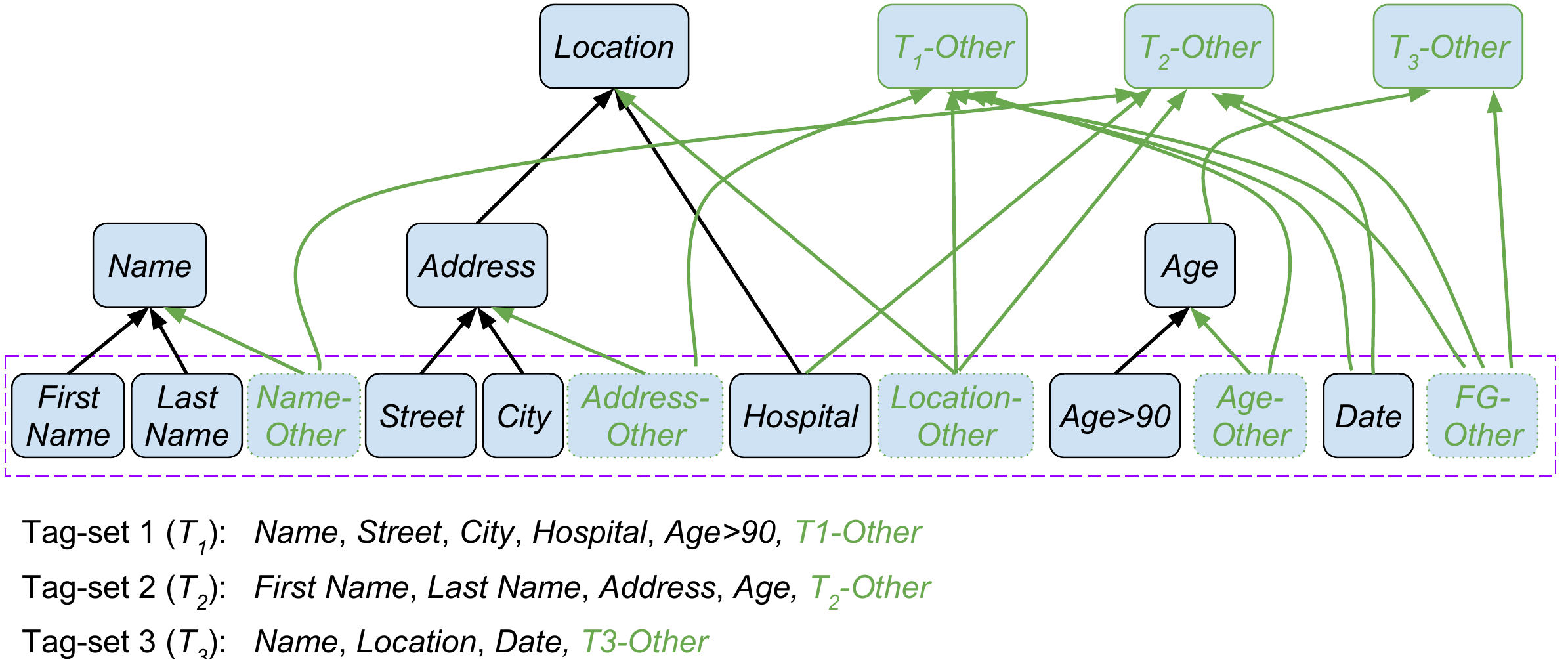}
\caption{The tag hierarchy in Fig.~\ref{fig:hierarchy} for three \tagsets{} after closure extension. Green nodes and edges were automatically added in this process. Fine-grained tags are surrounded by a dotted box.}
\label{fig:hierarchy_postprocessing}
\end{figure}


The models introduced in Sec.~\ref{ssec:base_adaptation} and \ref{ssec:multitasking} learn to predict a tag sequence for each training \tagset{} separately and they do not share parameters between tagging layers.
In addition, they require a post-processing step, outside of the model, for merging the tag sequences inferred for the different \tagsets{}.ß A simple concatenation of all training data is also not enough to accommodate the differences between the \tagsets{} within the model (see Sec.~\ref{sec:mtl_adaptation}). Moreover, none of these models utilizes the relations between tags, which are provided as input in the form of a tag hierarchy.

In this section, we propose a model that addresses these limitations. This model utilizes the given tag hierarchy at training time to learn a single, shared tagging layer that predicts only fine-grained tags. The hierarchy is then used during inference to map fine-grained tags onto a target \tagset{}.
Consequently, all tagging decisions are made in the model, without the need for a post-processing step.

\subsection{Notations}
\label{ssec:dag_notations}

In the input hierarchy DAG, each node represents some semantic role of words in sentences, (e.g. \qex{Name}). A directed edge $c \rightarrow d$ implies that $c$ is a hyponym of $d$, meaning $c$ captures a subset of the semantics of $d$. Examples include $\mathqex{LastName}\rightarrow\mathqex{Name}$, and $\mathqex{Street}\rightarrow\mathqex{Location}$ in Fig.~\ref{fig:hierarchy}. 
We denote the set of all tags that capture some subset of semantics of $d$ by $\mathrm{Sem}(d) = \{d\} \cup \{c | c \xrightarrow{\text{R}} d\}$, where $\xrightarrow{\text{R}}$ indicates that there is a directed path from $c$ to $d$ in the graph. For example, $\mathrm{Sem}(Name)=\{Name,LastName,FirstName\}$.

If a node $d$ has no hyponyms ($\mathrm{Sem}(d) = \{d\}$), it represents some fine-grained tag semantics. We denote the set of all fine-grained tags by \FGTS{}. We also denote all fine-grained tags that are hyponyms of $d$ by $\mathrm{Fine}(d) = \FGTSmath{} \cap \mathrm{Sem}(d)$, \eg{} $\mathrm{Fine}(Name)=\{LastName,FirstName\}$.
As mentioned above, our hierarchical model predicts tag sequences only from \FGTS{} and then maps them onto a target \tagset{}.

\subsection{Hierarchy extension with `Other' tags}
\label{ssec:preprocessing}

For each tag $d$ we would like the semantics captured by the union of semantics of all tags in $\mathrm{Fine}(d)$ to be exactly the semantics of $d$, making sure we will not miss any aspect of $d$ when predicting only over \FGTS{}. 
Yet, this semantics-equality property does not hold in general. One such example in Fig.~\ref{fig:hierarchy_postprocessing} is \qex{Age$>\!\!90$}$\rightarrow$\qex{Age}, because there may be age mentions below $90$ annotated in $T_2$'s dataset.

To fix the semantics-equality above, we use the notion of the \qex{Other} tag in NER, which has the semantics of ``\emph{all the rest}''. 
Specifically, for every $d \notin \FGTSmath{}$, a fine-grained tag \qex{$d$-Other} $\in \FGTSmath{}$ and an edge \qex{$d$-Other}$\rightarrow$\qex{$d$} are automatically added to the graph, hence \qex{$d$-Other}$\in \mathrm{Fine}(d)$. For instance, \qex{Age-Other}$\rightarrow$\qex{Age}. These new tags represent the aspects of $d$ not captured by the other tags in $\mathrm{Fine}(d)$.


Next a tag \qex{$T_i$-Other} is automatically added to each \tagset{} $T_i$, explicitly representing the ``\emph{all the rest}'' semantics of $T_i$. The labels for \qex{$T_i$-Other} are induced automatically from unlabeled tokens in the original $DS^r_i$ dataset.
To make sure that the semantics-equality property above also holds for \qex{$T_i$-Other}, a fine-grained tag \qex{FG-Other} is also added, which captures the ``\emph{all the rest}'' semantics at the fine-grained level. 
Then, each \qex{$T_i$-Other} is connected to all fine-grained tags that do not capture some semantics of the tags in $T_i$, defining:
\begin{align*}
\mathrm{Fine}(T_i\text{-Other}) = \FGTSmath\ \ \setminus \bigcup_{d \in T_i \smallsetminus\!\{T_i\text{-Other}\}}\!\!\!\mathrm{Sem(d)}
\end{align*}
This mapping is important at training time, where \qex{$T_i$-Other} labels are used as distant supervision over their related fine-grained tags (Sec. \ref{ssec:tag_hier_model}).
Fig. \ref{fig:hierarchy_postprocessing} depicts our hierarchy example after this step. We emphasize that all extensions in this step are done automatically as part of the model's algorithm.


\subsection{NER model with tag hierarchy}
\label{ssec:tag_hier_model}
One outcome of the extension step is that the set of fine-grained tags \FGTS{} covers all distinct fine-grained semantics across all \tagsets{}.
In the following, we train a single NER model (Sec.~\ref{ssec:base_model}) that predicts sequences of tags from the \FGTS{} \tagset{}. As there is only one tagging layer, model parameters are shared across all training examples.


At inference time, this model predicts the most likely fine-grained tag sequence $y^{fg}$ for the input $x$.
As the model outputs only a single sequence, post-processing consolidation is not needed.
The tag hierarchy is used to map each predicated fine-grained tag $y^{fg}_i$ to a tag in a test \tagset{} $T^s$ by traversing the out-edges of $y^{fg}_i$ until a tag in $T^s$ is reached. This procedure is also used in the baseline models (see Sec.~\ref{ssec:base_adaptation}) for mapping their predictions onto the test \tagset{}. However, unlike the baselines, which end with multiple candidate predictions in the test \tagset{} and need to consolidate between them, here, only a single fine-grained tag sequence is mapped, so no further consolidation is needed.

At training time, each example $x$ that belongs to some training dataset $DS^r_i$ is labeled with a gold-standard tag sequence $y$ where the tags are taken only from the corresponding \tagset{} $T^r_i$. 
This means that tags $\{y_i\}$ are not necessarily fine-grained tags, so there is no direct supervision for predicting fine-grained tag sequences. 
However, each gold label $y_i$ provides distant supervision over its related fine-grained tags, $\mathrm{Fine}(y_i)$. It indicates that one of them is the correct fine-grained label without explicitly stating which one, so we consider all possibilities in a probabilistic manner. 

Henceforth, we say that a fine-grained tag sequence $y^{fg}$ \emph{agrees with} $y$ if $\bigforall_{i}\ y^{fg}_i \in \mathrm{Fine}(y_i)$, \ie{} $y^{fg}$ is a plausible interpretation for $y$ at the fine-grained tag level. 
For example, following Fig.~\ref{fig:hierarchy_postprocessing}, sequences [\qex{Hospital}, \qex{City}] and [\qex{Street}, \qex{City}] agree with [\qex{Location}, \qex{Location}], unlike [\qex{City}, \qex{Last Name}].
We denote all fine-grained tag sequences that agree with $y$ by $\mathrm{AgreeWith}(y)$.

Using this definition, the tag-hierarchy model is trained with the loss function:
\begin{align}
loss(y)& = -log(\frac{Z_y}{Z}) \\
Z_y & = \sum_{y^{fg} \in \mathrm{AgreeWith}(y)} \phi({y^{fg}}) \\
Z & = \sum_{y^{fg}} \phi({y^{fg}})
\end{align}
where $\phi(y)$ stands for the model's score for sequence $y$, viewed as unnormalized probability. $Z$ is the standard CRF partition function over all possible fine-grained tag sequences. $Z_y$, on the other hand, accumulates scores only of fine-grained tag sequences that agree with $y$. Thus, this loss function aims at increasing the summed probability of all fine-grained sequences agreeing with $y$.
Both $Z_y$ and $Z$ can be computed efficiently using the Forward-Backward algorithm \cite{CRF}.

We note that we also considered finding the most likely tag sequence over a test \tagset{} at inference time by summing the probabilities of all fine-grained tag sequences that agree with each candidate sequence $y$:
$
\max_{y}\sum_{y^{fg} \in \mathrm{AgreeWith}(y)} \phi({y^{fg}})
$.
However, this problem is NP-hard \cite{Lyngso2002}. We plan to explore other alternatives in future work.




\section{Experimental Settings}
\label{sec:settings}

\begin{table}
\centering
\scriptsize
\begin{tabular}{ |l|r|r|r| } 
 \hline
 \multicolumn{1}{|c|}{\multirow{2}{*}{Dataset}} & \multicolumn{1}{c|}{\Tagset{}} & \multicolumn{1}{c|}{\multirow{2}{*}{\# Tokens}} & \multicolumn{1}{l|}{Tagged \multirow{2}{*}{(\%)}} \\
 & \multicolumn{1}{c|}{Size} & & \multicolumn{1}{l|}{Tokens} \\
 \hline
 \hline
 \ISIX{} (train) & \multirow{2}{*}{7} & 387,126 & 4.6 \\ 
 \ISIX{} (test) & & 163,488 & 4.2 \\ 
 \hline
 \IFOURTEEN{} (train) & \multirow{3}{*}{17} & 336,422 & 4.4 \\ 
 \IFOURTEEN{} (dev) & & 152,895 & 5.0 \\
 \IFOURTEEN{} (test) & & 316,212 & 4.6 \\ 
 \hline
 \PHYS{} (test) & 6 & 335,383 & 0.7 \\
 \hline
 \CONLL{} (train) & \multirow{3}{*}{4} & 203,621 & 16.7 \\
 \CONLL{} (dev) & & 51,362 & 16.7 \\
 \CONLL{} (test) & & 46,435 & 18.1 \\
 \hline
 \ONTO{} (train) & \multirow{2}{*}{18} & 1,304,491 & 13.1 \\
 \ONTO{} (test) & & 162,971 & 14.2 \\
 \hline
\end{tabular}
\caption{Dataset statistics. \emph{Tokens tagged} refer to percentage of tokens tagged not as \qex{Other}.}
\label{tab:datasets}
\end{table}

To test the tag-hierarchy model under \HETTAG{} scenarios, we conducted experiments using datasets from two domains. We next describe these datasets as well as implementation details for the tested models. Sec.~\ref{sec:results} then details the experiments and their results.

\subsection{Datasets}
\label{ssec:datasets}







Five datasets from two domains, \emph{medical} and \emph{news}, were used in our experiments. Table~\ref{tab:datasets} summarizes their main statistics. 

For the medical domain we used the datasets
I2B2-2006 (denoted \ISIX{}) \cite{I2B206}, I2B2-2014 (denoted \IFOURTEEN{}) \cite{I2B214} and the PhysioNet golden set (denoted \PHYS{}) \cite{PhysioNet}. 
These datasets are all annotated for the NER task of de-identification (a.k.a text anonymization) \cite{Dernoncourt2017}. Still, as seen in Table~\ref{tab:datasets}, each dataset is annotated with a different \tagset{}.
Both \ISIX{} and \IFOURTEEN{} include train and test sets, while \PHYS{} contains only a test set.

For the news domain we used the English part of CONLL-2003 (denoted \CONLL{}) \cite{CONLL2003} and OntoNotes-v5 (denoted \ONTO{}) \cite{OntoNotes}, both with train and test sets.  We note that \IFOURTEEN{}, \CONLL{} and \ONTO{} also contain a dev-set, which is used for hyper-param tuning (see below).

In all experiments, each example is a full document. Each document is split into tokens on white-spaces and punctuation. A tag-hierarchy covering the 57 tags from all five datasets was given as input to all models in all experiments. We constructed this hierarchy manually. The only non-trivial tag was \qex{Location}, which in \IFOURTEEN{} is split into finer tags (\qex{City}, \qex{Street} etc.) and includes also hospital mentions in \CONLL{} and \ONTO{}. We resolved these relations similarly to the graph in Figure~\ref{fig:hierarchy}.


\subsection{Compared Models}
\label{ssec:compared_models}




Four models were compared in our experiments:

\textbf{\ConcatM{}}\ \ \ A single NER model on the concatenation of datasets and \tagsets{} (Sec.~\ref{sec:mtl_adaptation}).

\textbf{\IndM{}}\ \ \ Combining predictions of independent NER models, one per \tagset{} (Sec.~\ref{ssec:base_adaptation}).

\textbf{\MtlM{}}\ Multitasking over training \tagsets{} (Sec.~\ref{ssec:multitasking}).

\textbf{\HierM{}}\ \ \ A tag hierarchy employed within a single base model (Sec.~\ref{sec:model}).

\ 

All models are based on the neural network described in Sec.~\ref{ssec:base_model}. We tuned the hyper-params in the base model to achieve \SOTA{} results for a single NER model on \CONLL{} and \IFOURTEEN{} when trained and tested on the same dataset \cite{Strubell2017,Dernoncourt2017} (see Table~\ref{tab:F1_upper}). This is done to maintain a constant baseline, and is also due to the fact that \ISIX{} does not have a standard dev-set. 

We tuned hyper-params over the dev-sets of \CONLL{} and \IFOURTEEN{}. For character-based embedding we used a single bidirectional LSTM \cite{LSTM} with hidden state size of 25. For word embeddings we used pre-trained GloVe embeddings\footnote{\url{nlp.stanford.edu/data/glove.6B.zip}} \cite{Glove}, without further training. For token recoding we used a two-level stacked bidirectional LSTM \cite{Graves2013} with both output and hidden state of size 100.

Once these hyper-params were set, no further tuning was made in our experiments, which means all models for \HETTAGS{} were tested under the above fixed hyper-param set. In each experiment, each model was trained until convergence on the respective training set.



\begin{table}[t]
\centering
\scriptsize
\begin{tabular}{ |l|r|r|r|r| } 
 \hline
& \multicolumn{1}{c|}{\ISIX{}} & \multicolumn{1}{c|}{\IFOURTEEN{}} & \multicolumn{1}{c|}{\CONLL{}} & \multicolumn{1}{c|}{\ONTO{}} \\
 \hline
 Micro avg. F1 & 0.894 & 0.960 & 0.926 & 0.896 \\ 
 \hline
\end{tabular}
\caption{F1 for training and testing a single base NER model on the same dataset.}
\label{tab:F1_upper}
\end{table}

\begin{table}[t]
\centering
\scriptsize
\begin{tabular}{ |l|r|r|r|r| } 
 \hline
& \multicolumn{4}{c|}{Tag Frequency in training / test (\%)} \\
& \multicolumn{1}{c|}{\ISIX{}} & \multicolumn{1}{c|}{\IFOURTEEN{}} & \multicolumn{1}{c|}{\CONLL{}} & \multicolumn{1}{c|}{\ONTO{}} \\
 \hline
 \hline
 Name & 1.4 / 1.3 & 1.0 / 1.0 & 4.3 / 4.9 & 3.1 / 2.9 \\ 
 \hline
 Date & 1.7 / 1.5 & 2.4 / 2.5 & 0 / 0 & 2.7 / 3.1 \\ 
 \hline
 Location & 0.1 / 0.1 & 0.2 / 0.3 & 3.2 / 3.4 & 2.7 / 3.2 \\
 \hline
 Hospital & 0.6 / 0.7 & 0.3 / 0.3 & 0 / 0 & 0 / 0 \\
 \hline
\end{tabular}
\caption{Occurrence statistics for tags used in the \tagset{} extension experiment, reported as \% out of all tokens in the training and test sets of each dataset.}
\label{tab:freq_tags}
\end{table}

\begin{table}[t]
\centering
\scriptsize
\begin{tabular}{ |ll@{\hskip1pt}|c|c|c@{\hskip1pt}| } 
 \hline
F1 AVERAGE & & \multicolumn{3}{|c|}{Model} \\
\multicolumn{1}{|l}{Extending Tag} & \multicolumn{1}{l}{Base Dataset} & \multicolumn{1}{|c|}{\bf Hier} & \multicolumn{1}{c|}{\bf Indep} & \multicolumn{1}{c|}{\bf MTL} \\
 \hline
 \hline
 Date & \IFOURTEEN{} & \textbf{0.806} & 0.795 & 0.787 \\
 & \ISIX{} & 0.756 & 0.761 & \textbf{0.787} \\
 & \ONTO{} & \textbf{0.835} & 0.828 & 0.819 \\
 \hline
 Date Total & & \textbf{0.799} & 0.795 & 0.798 \\
 \hline
 \hline
  Hospital & \IFOURTEEN{} & 0.931 & \textbf{0.941} & 0.918 \\
  & \ISIX{} & \textbf{0.867} & 0.866 & 0.853 \\
  \hline
  Hospital Total & & 0.899 & \textbf{0.904} & 0.885 \\
  \hline
  \hline
 Location & \CONLL{} & \textbf{0.801} & 0.784 & 0.793 \\
 & \IFOURTEEN{} & \textbf{0.953} & 0.913 & 0.905 \\
 & \ISIX{} & \textbf{0.877} & 0.848 & 0.820 \\
 & \ONTO{} & \textbf{0.785} & 0.694 & 0.692 \\
 \hline
 Location Total & & \textbf{0.854} & 0.810 & 0.802 \\
 \hline
 \hline
 Name & \CONLL{} & \textbf{0.847} & 0.759 & 0.729 \\
 & \IFOURTEEN{} & \textbf{0.918} & 0.880 & 0.902 \\
 & \ISIX{} & 0.740 & \textbf{0.743} & 0.729 \\
 & \ONTO{} & \textbf{0.878} & 0.862 & 0.862 \\
 \hline
 Name Total & & \textbf{0.846} & 0.811 & 0.806 \\
 \hline
 \hline
 \textbf{Grand Total} & & \textbf{0.854} & 0.823 & 0.816 \\
\hline
\end{tabular}
\caption{F1 in the \tagset{} extension experiment, averaged over extending datasets for every base dataset.}
\label{tab:extension_results}
\end{table}

\begin{table}[t]
\centering
\scriptsize
\begin{tabular}{ |lll@{\hskip1pt}|r|r|r@{\hskip1pt}| } 
 \hline
F1 & & & \multicolumn{3}{c|}{Model} \\
\multicolumn{1}{|l}{Tag} & \multicolumn{1}{l}{Base} & \multicolumn{1}{l|}{Extending} & \multicolumn{1}{c|}{\bf Hier} & \multicolumn{1}{c|}{\bf Indep} & \multicolumn{1}{c|}{\bf MTL} \\
 \hline
 \hline
 Date & \IFOURTEEN{} & \ISIX{} & 0.899 & & *\textbf{0.903} \\
 & & \ONTO{} & *\textbf{0.713} & 0.686 & 0.671 \\
 & \ISIX{} & \ONTO{} & 0.641 & *\textbf{0.681} & \\
  & \ONTO{} & \ISIX{} & *\textbf{0.834} & & 0.807 \\
  \hline
  \hline
 Location & \CONLL{} & \IFOURTEEN{} & *\textbf{0.818} & 0.783 & \\
 & & \ISIX{} & *\textbf{0.748} & & 0.730 \\
 & & \ONTO{} & *\textbf{0.836} & 0.830 & \\
 & \IFOURTEEN{} & \CONLL{} & *\textbf{0.954} & 0.899 & 0.887 \\
 & & \ONTO{} & *\textbf{0.951} & 0.921 & 0.907 \\
 & \ISIX{} & \CONLL{} & \textbf{0.876} & 0.816 & 0.760 \\
  & & \ONTO{} & *\textbf{0.869} & 0.847 & 0.812 \\
  & \ONTO{} & \CONLL{} & *\textbf{0.747} & 0.701 & 0.703 \\
  & & \IFOURTEEN{} & \textbf{0.793} & 0.691 & 0.707 \\
  & & \ISIX{} & *\textbf{0.814} & 0.691 & \\
  \hline
  \hline
  Name & \CONLL{} & \IFOURTEEN{} & *\textbf{0.855} & & 0.690 \\
  & & \ISIX{} & *\textbf{0.827} & 0.666 & 0.631 \\
  & & \ONTO{} & \textbf{0.860} & 0.841 & \\
  & \IFOURTEEN{} & \CONLL{} & *\textbf{0.900} & 0.863 & \\
  & & \ISIX{} & *\textbf{0.943} & 0.893 & \\
  & & \ONTO{} & *\textbf{0.911} & 0.882 & 0.891 \\
  & \ISIX{} & \CONLL{} & *\textbf{0.662} & & 0.653 \\
  & \ONTO{} & \CONLL{} & *\textbf{0.895} & 0.888 & \\
  & & \IFOURTEEN{} & *\textbf{0.892} & 0.872 & \\
  & & \ISIX{} & *\textbf{0.846} & 0.827 & \\
  \hline
 
\hline
\end{tabular}
\caption{F1 for \tagset{} extensions with more than 100 collisions. Blank entries indicate fewer than 100 collisions. (*) indicates all results that are statistically significantly better than others in that row.}
\label{tab:extension_collisions}
\end{table}

\section{Experiments and Results}
\label{sec:results}

We performed two experiments. The first refers to selective annotation, in which an existing \tagset{} is extended with a new tag by annotating a new dataset only with the new tag.
The second experiment tests the ability of each model to integrate two full \tagsets{}.

In all experiments we assess model performance via micro-averaged tag F1, in accordance with CoNLL evaluation \cite{CONLL2003}. Statistical significance was computed using the Wilcoxon two-sided signed ranks test at $p=0.01$ \cite{Wilcoxon1945}.
We next detail each experiment and its results.

In all our experiments, we found the performance of the different consolidation methods (Sec.~\ref{ssec:base_adaptation}) to be on par. One reason that using model scores does not beat random selection may be due to the overconfidence of the tagging models -- their prediction probabilities are close to 0 or 1.
We report figures for random selection as representative of all consolidation methods.

\subsection{\Tagset{} extension experiment} \label{ExtensionExperiment}

In this experiment, we considered the 4 most frequent tags that occur in at least two of our datasets: \qex{Name}, \qex{Date}, \qex{Location} and \qex{Hospital} (Table~\ref{tab:freq_tags} summarizes their statistics). 
For each frequent tag $t$ and an ordered pair of datasets in which $t$ occurs, we constructed new training sets by removing $t$ from the first training set (termed \emph{base dataset}) and remove all tags but $t$ from the second training set (termed \emph{extending dataset}). For example, for the \textit{triplet} of \{ \qex{Name}, \IFOURTEEN{}, \ISIX{}\}, we constructed a version of \IFOURTEEN{} without \qex{Name} annotations and a version of \ISIX{} containing only annotations for \qex{Name}. This process yielded 32 such triplets.

For every triplet, we train all tested models on the two modified training sets and test them on the test-set of the base dataset (\IFOURTEEN{} in the example above). Each test-set was not altered and contains all tags of the base \tagset{}, including $t$.

\ConcatM{} performed poorly in this experiment. For example, on the dataset extending \IFOURTEEN{} with \qex{Name} from \ISIX{}, \ConcatM{} tagged only one \qex{Name} out of over 4000 \qex{Name} mentions in the test set. Given this, we do not provide further details of the results of \ConcatM{} in this experiment.

For the three models tested, this experiment yields 96 results. The main results\footnote{Detailed results for all 96 tests are given in the Appendix.} of this experiment are shown in Table~\ref{tab:extension_results}. Surprisingly, in more tests \IndM{} outperformed \MtlM{} than vice versa, adding to prior observations that multitasking can hurt performance instead of improving it \cite{Bingel2017,Alonso2017,Bjerva2017}. But, applying a shared tagging layer on top of a shared text representation boosts the model's capability and stability. Indeed, overall, \HierM{} outperforms the other models in most tests, and in the rest it is similar to the best performing model. 

Analyzing the results, we noticed that the gap between model performance increases when more collisions are encountered for \MtlM{} and \IndM{} at post-processing time (see Sec.~\ref{ssec:base_adaptation}). The amount of collisions may be viewed as a predictor for the baselines' difficulty to handle a specific \HETTAGS{} setting. Table~\ref{tab:extension_collisions} presents the tests in which more than 100 collisions were detected for either \IndM{} or \MtlM{}, constituting 66\% of all test triplets. In these tests, \HierM{} is a clear winner, outperforming the compared models in all but two comparisons, often by a significant margin. 

Finally, we compared the models trained with selective annotation to an ``upper-bound'' of training and testing a single NER model on the same dataset with all tags annotated (Table~\ref{tab:F1_upper}). As expected, performance is usually lower with selective annotation. But, the drop intensifies when the base and extending datasets are from different domains -- medical and news. In these cases, we observed that \HierM{} is more robust. Its drop compared to combining datasets from the same domain is the least in almost all such combinations. Table~\ref{tab:cross_domain} provides some illustrative examples.

\begin{table}[t]
\centering
\scriptsize
\begin{tabular}{ |lll@{\hskip1pt}|r|r|r@{\hskip1pt}| } 
 \hline
F1 & & & \multicolumn{3}{c|}{Model} \\
\multicolumn{1}{|l}{Tag} & \multicolumn{1}{l}{Base} & \multicolumn{1}{l|}{Extending} & \multicolumn{1}{c|}{\bf Hier} & \multicolumn{1}{c|}{\bf Indep} & \multicolumn{1}{c|}{\bf MTL} \\
 \hline
 \hline
 \multirow{2}{*}{Location} & \multirow{2}{*}{\IFOURTEEN{}} & \ISIX{} & 0.953 & 0.919 & 0.919 \\
 & & \ONTO{} & 0.954 & 0.899 & 0.887 \\
 \hline
 \multirow{2}{*}{Name} & \multirow{2}{*}{\CONLL{}} & \ISIX{} & 0.846 & 0.827 & 0.809 \\
 & & \ONTO{} & 0.895 & 0.888 & 0.890 \\
 \hline
\end{tabular}
\caption{Examples for performance differences when base datasets are extended with an in-domain dataset compared to an out-of-domain dataset.}
\label{tab:cross_domain}
\end{table}

\subsection{Full \tagset{} integration experiment}

\begin{table}[t]
\centering
\scriptsize
\begin{tabular}{ |l|r|r|r| } 
 \hline
 \multicolumn{1}{|@{\hskip1pt}l|}{F1} & \multicolumn{3}{c|}{Test Set} \\
 \multicolumn{1}{|c|}{Model} & \ISIX{} & \IFOURTEEN{} & \PHYS{} \\
 \hline
 \ISIX{} &  *0.894 & 0.730 & 0.637 \\
 \IFOURTEEN{} & 0.714 & *\textbf{0.960} & 0.712 \\
 \hline
 \ConcatM{} & 0.827 & 0.809 & 0.621 \\
 \IndM{} & 0.760 & 0.861 & 0.640 \\
 \MtlM{} & 0.81 & 0.862 & *0.739 \\
 \HierM{} & *\textbf{0.900} & *0.958 & *\textbf{0.760} \\
 \hline
 \hline
 Collisions & \multicolumn{3}{c|}{Test Set} \\
  & \ISIX{} & \IFOURTEEN{} & \PHYS{} \\
 \hline
 \IndM{} & 224 & 1272 & 114 \\
 \MtlM{} & 158 & 584 & 44 \\
 \hline
\end{tabular}
\caption{F1 for combining \ISIX{} and \IFOURTEEN{}. The top two models were trained only on a single dataset. The lower table part holds the number of collisions at post-processing. (*) indicates results that are statistically significantly better than others in that column.}
\label{tab:full_combination}
\end{table}

A scenario distinct from selective annotation is the integration of full \tagsets{}. On one hand, more training data is available for similar tags. On the other hand, more tags need to be consolidated among the \tagsets{}.

To test this scenario, we trained the tested model types on the training sets of \ISIX{} and \IFOURTEEN{}, which have different \tagsets{}. The models were evaluated both on the test sets of these datasets and on \PHYS{}, an unseen test-set that requires the combination of the two training \tagsets{} for full coverage of its \tagset{}. We also compared the models to single models trained on each of the training sets alone.
Table~\ref{tab:full_combination} displays the results. 

As expected, single models do well on the test-set companion of their training-set but they under-perform on the other test-sets. This is expected because the \tagset{} on which they were trained does not cover well the \tagsets{} in the other test-sets.

When compared with the best-performing single model, using \ConcatM{} shows reduced results on all 3 test sets. This can be attributed to reduced performance for types that are semantically different between datasets (e.g. \qex{Date}), while performance on similar tags (e.g. \qex{Name}) does not drop. 

Combining the two training sets using either \IndM{} or \MtlM{} leads to substantial performance drop in 5 out of 6 test-sets compared to the best-performing single model. This is strongly correlated with the number of collisions encountered (see Table~\ref{tab:full_combination}). Indeed, the only competitive result, \MtlM{} tested on \PHYS{}, had less than 100 collisions. This demonstrates the non triviality in real-world \tagset{} integration, and the difficulty of resolving tagging decisions across \tagsets{}.

By contrast, \HierM{} has no performance drop compared to the single models trained and tested on the same dataset.
Moreover, it is the best performing model on the unseen \PHYS{} test-set, with 6\% relative improvement in F1 over the best single model.
This experiment points up the robustness of the tag hierarchy approach 
when applied to this \HETTAG{} scenario.



\section {Related Work}
\label{sec:related}








\citet{Collobert2011} introduced the first competitive NN-based NER that required little or no feature engineering.
\citet{Huang2015NNbasedNER} combined LSTM with CRF, showing performance similar to non-NN models. \citet{Lample2016NNbasedNER} extended this model with character-based embeddings in addition to word embedding, achieving \SOTA{} results.
Similar architectures, such as combinations of convolutional networks as replacements of RNNs were shown to out-perform previous NER models \cite{Ma2016,Chiu2016,Strubell2017}.

\citet{Dernoncourt2017} and \citet{Liu2017deid} showed that the LSTM-CRF model achieves \SOTA{} results also for de-identification in the medical domain. \citet{Lee2017} demonstrated how performance drops significantly when the LSTM-CRF model is tested under transfer learning within the same domain in this task.

\citet{Collobert2008} introduced MTL for NN, and other works followed, showing it helps in various NLP tasks \cite{Chen2016,Peng2017}. \citet{Sogaard2016} and \citet{Hashimoto2017} argue that cascading architectures can improve MTL performance. Several works have explored conditions for successful application of MTL \cite{Bingel2017,Bjerva2017,Alonso2017}.


Few works attempt to share information across datasets at the tagging level. \citet{greenberg2018} proposed a single CRF model for tagging with \HETTAGS\ but without a hierarchy. They show the utility of this method for in-domain datasets with a balanced tag distribution. Our model can be viewed as an extension of theirs for tag hierarchies. \citet{Augenstein2018} use tag embeddings in MTL to further propagate information between tasks. \citet{Li2017} propose to use a \tagset{} made of cross-product of two different POS \tagsets{} and train a model for it. Given the explosion in \tagset{} size, they introduce automatic pruning of cross-product tags. 
\citet{Kim2015} and \citet{Qu2016} automatically learn correlations between \tagsets{}, given training data for both \tagsets{}. They rely on similar contexts for related source and target tags, such as \qex{professor} and \qex{student}.


Our tag-hierarchy model was inspired by recent work on hierarchical multi-label classification \cite{silla2011survey,zhang2014review}, and can be viewed as an extension of this direction onto sequences tagging.

\section{Conclusions}
\label{sec:conclusions}

We proposed a tag-hierarchy model for the \HETTAGS{} NER setting, which does not require a consolidation post-processing stage.
In the conducted experiments, the proposed model consistently outperformed the baselines in difficult tagging cases and showed robustness when applying a single trained model to varied test sets.

In the case of integrating datasets from the news and medical domains we found the blending task to be difficult.
In future work, we'd like to improve this integration in order to gain from training on examples from different domains for tags like \qex{Name} and \qex{Location}. 

\section*{Acknowledgments}
The authors would like to thank Yossi Matias, Katherine Chou, Greg Corrado, Avinatan Hassidim, Rony Amira, Itay Laish and Amit Markel for their help in creating this work.

\bibliography{hetero_bibliography.bib}
\bibliographystyle{acl_natbib}

\newpage
\appendix
\section{Experiment Results}
Full experiment results for Section \ref{ExtensionExperiment}
\begin{table}[h]
\centering
\scriptsize
\begin{tabular}{ |lll@{\hskip1pt}|r|r|r@{\hskip1pt}| } 
 \hline
F1 & & & \multicolumn{3}{c|}{Model} \\
\multicolumn{1}{|l}{Tag} & \multicolumn{1}{l}{Base} & \multicolumn{1}{l|}{Extending} & \multicolumn{1}{c|}{\bf Hier} & \multicolumn{1}{c|}{\bf Indep} & \multicolumn{1}{c|}{\bf MTL} \\
 \hline
 \hline
 Date & \IFOURTEEN{} & \ISIX{} & 0.899 & 0.904 & 0.903 \\
 &  & \ONTO{} & 0.713 & 0.686 & 0.671 \\
 & \ISIX{} & \IFOURTEEN{} & 0.871 & 0.840 & 0.875 \\
 &  & \ONTO{} & 0.641 & 0.681 & 0.698 \\
 & \ONTO{} & \IFOURTEEN{} & 0.837 & 0.830 & 0.831 \\
  &  & \ISIX{} & 0.834 & 0.826 & 0.807 \\
  \hline
  \hline
 Hospital & \IFOURTEEN{} & \ISIX{} & 0.931 & 0.941 & 0.918 \\
 & \ISIX{} & \IFOURTEEN{} & 0.867 & 0.866 & 0.853 \\
  \hline
  \hline
 Location & \CONLL{} & \IFOURTEEN{} & 0.818 & 0.783 & 0.812 \\
 & & \ISIX{} & 0.748 & 0.739 & 0.730 \\
 & & \ONTO{} & 0.836 & 0.830 & 0.836 \\
 & \IFOURTEEN{} & \CONLL{} & 0.954 & 0.899 & 0.887 \\
 & & \ISIX{} & 0.953 & 0.919 & 0.919 \\
 & & \ONTO{} & 0.951 & 0.921 & 0.907 \\
 & \ISIX{} & \CONLL{} & 0.876 & 0.816 & 0.760 \\
 & & \IFOURTEEN{} & 0.886 & 0.883 & 0.888 \\
  & & \ONTO{} & 0.869 & 0.847 & 0.812 \\
  & \ONTO{} & \CONLL{} & 0.747 & 0.701 & 0.703 \\
  & & \IFOURTEEN{} & 0.793 & 0.691 & 0.707 \\
  & & \ISIX{} & 0.814 & 0.691 & 0.666 \\
  \hline
  \hline
  Name & \CONLL{} & \IFOURTEEN{} & 0.855 & 0.771 & 0.690 \\
  & & \ISIX{} & 0.827 & 0.666 & 0.631 \\
  & & \ONTO{} & 0.860 & 0.841 & 0.867 \\
  & \IFOURTEEN{} & \CONLL{} & 0.900 & 0.863 & 0.890 \\
  & & \ISIX{} & 0.943 & 0.893 & 0.927 \\
  & & \ONTO{} & 0.911 & 0.882 & 0.891 \\
  & \ISIX{} & \CONLL{} & 0.662 & 0.679 & 0.653 \\
  & & \IFOURTEEN{} & 0.834 & 0.824 & 0.808 \\
  & & \ONTO{} & 0.726 & 0.726 & 0.727 \\
  & \ONTO{} & \CONLL{} & 0.895 & 0.888 & 0.890 \\
  & & \IFOURTEEN{} & 0.892 & 0.872 & 0.886 \\
  & & \ISIX{} & 0.846 & 0.827 & 0.809 \\
  \hline
 
\hline
\end{tabular}
\end{table}
\end{document}